# BIRNet: Brain Image Registration Using Dual-Supervised Fully Convolutional Networks

Jingfan Fan[1], Xiaohuan Cao[1,2], Pew-Thian Yap[1], and Dinggang Shen[1,3*]

[1] Department of Radiology and BRIC, University of North Carolina at Chapel Hill, Chapel Hill, NC, USA

[2] School of Automation, Northwestern Polytechnical University, Xi'an, China

[3] Department of Brain and Cognitive Engineering, Korea University, Seoul 02841, Republic of Korea

[*] Corresponding author: dgshen@med.unc.edu

**Abstract.** In this paper, we propose a deep learning approach for image registration by predicting deformation from image appearance. Since obtaining ground-truth deformation fields for training can be challenging, we design a fully convolutional network that is subject to dual-guidance: (1) Coarse guidance using deformation fields obtained by an existing registration method; and (2) Fine guidance using image similarity. The latter guidance helps avoid overly relying on the supervision from the training deformation fields, which could be inaccurate. For effective training, we further improve the deep convolutional network with gap filling, hierarchical loss, and multi-source strategies. Experiments on a variety of datasets show promising registration accuracy and efficiency compared with state-of-the-art methods.

**Keywords:** Image registration; Convolutional neural networks; Brain MR image; Hierarchical registration.

## 1. Introduction

Deformable registration establishes anatomical correspondences between a pair of images. Although many registration algorithms have been proposed in the past decades, registration is still a challenging problem since it often involves computationally expensive high-dimensional optimization and task-dependent parameter tuning. Besides, although deep learning techniques have already shown high performance in many medical image analysis tasks, such as segmentation (Ronneberger et al., 2015) or classification (He et al., 2015), it is still hard to directly solve the registration problem due to the lack of the ideal ground-truth deformations, which are difficult to manually annotate in practice.

In this paper, we present a brain image registration network (BIRNet) for learning-based deformable registration. We will introduce a novel *hierarchical dual-supervised fully convolutional neural network* (FCN) to deal with the lack of ground truth for training. BIRNet predicts the deformation field in one-pass and is insensitive to parameter tuning. Our main contributions are summarized below:

1. We propose a *dual-supervised* deep learning strategy that involves dual-guidance: 1) Coarse guidance using the deformation field estimated by an existing registration method, and 2) Fine guidance based on the similarity between the template and the warped subject image given the estimated deformation. The latter guidance helps avoid overly relying on the supervision from the training deformation fields, which could be inaccurate.
2. To improve efficiency, U-Net (Ronneberger et al., 2015) will be used as the basic architecture. For accuracy, we further propose to use *gap filling* for learning more high-level features and use *multi-channel inputs* (i.e., the gradient map and difference map) for better informing the registration network.
3. We propose an end-to-end framework for deformation prediction in one-pass without the need for parameter tuning.

We validate our method on a variety of datasets and registration tasks. Experimental results confirm the accuracy and robustness of the proposed method.

The remaining part of this paper is organized as follow. Section 2 reviews related works. Section 3 details the proposed method, including an overview (Section 3.1), the network design (Section 3.2), and dataset augmentation (Section 3.3). Section 4 presents experimental results, and Section 5 discusses future directions and applications.

## 2. Related Works

### 2.1 Registration via Optimization

The optimization based deformable registration methods can be divided into two categories (Oliveira and Tavares, 2014; Sotiras et al., 2013): intensity-based (Johnson and Christensen, 2002; Klein et al., 2010; Myronenko and Song, 2010; Vercauteren et al., 2009) and feature-based (Auzias et al., 2011; Avants et al., 2008; Ou et al., 2011; Shen and Davatzikos, 2002; Wu et al., 2014; Wu et al., 2010). The deformable registration is often based on linear (rigid/affine) registration (Fan et al., 2016a; Fan et al., 2016b; Fan et al., 2017), where the linear registration intends to globally align the two images and the deformation registration is used to correct the local deformations. But unlike linear registration, deformable registration is an often ill-posed high-dimensional optimization problem. Therefore, most of them involve time-consuming iterative optimization and task-sensitive parameter tuning.

Implementation using graphics processing units (GPUs) has becoming more common for improving computational efficiency. Voxel- or patch-level computation, such as interpolation and local similarity, can be parallelized and accelerated significantly (Fluck et al., 2011), often increasing the speed by a factor of more than 10 (Samant et al., 2015; Shamonin et al., 2013; ur Rehman et al., 2009). However, not all processes can be accelerated, especially those involving iterative optimization and huge memory swapping between CPUs and GPUs (Yang et al., 2017). Moreover, a significant amount of effort is often needed to redesign and port algorithms for GPUs (Fluck et al., 2011).

### 2.2 Registration via Learning

Learning-based statistical models have been widely investigated to improve registration performance by establishing the correlation between the deformation field and images (under registration) based on a training dataset. Learning-based registration methods predict deformation parameters by using machine learning algorithms, such as principal components analysis (Loeckx et al., 2003; Rueckert et al., 2001, 2003), support vector regression (Kim et al., 2012), sparse representation (Kim et al., 2015; Wang et al., 2015), semi-coupled dictionary (Cao et al., 2015), and gradient boosted trees (Gutiérrez-Becker et al., 2016; Gutierrez-Becker et al., 2017). For example, *Kim et al*. (Kim et al., 2015) and *Wang et al*. (Wang et al., 2015) proposed to predict the deformations of a number of distinctive key points in the brain. *Gutierrez et al*. (Gutiérrez-Becker et al., 2016; Gutierrez-Becker et al., 2017) proposed to predict deformation parameters via a regression model based on gradient boosted trees, instead of directly minimizing a registration energy.

### 2.3 Registration via Deep Learning

More recently, deep learning methods such as convolutional neural networks (CNN) has been shown to be applicable for registration. *Soot et al.* (Sokooti et al., 2017) proposed RegNet to estimate the displacement vector field for a pair of chest CT images. *Cao et al.* (Cao et al., 2017) used an equalized active-points sampling strategy to build a similarity-steered CNN model to predict the deformations associated with the active points. *Yang et al.* (Yang et al., 2017) predicted the momenta of the deformation in a large deformation diffeomorphic metric mapping (LDDMM) setting. *Rohe et al.* (Rohé et al., 2017) built reference deformations for training by registering manually delineated regions of interest (ROIs). All the recent deep learning based registration methods have to spend time on carefully building the reference deformations due to the lack of the ideal ground-truth deformations for training.

## 3. Method

### 3.1 Overview

The goal of image registration is to determine a deformation field $\phi$ that warps a subject image $S \in \mathbb{R}^3$ to a template image $T \in \mathbb{R}^3$, so that the warped image $S \circ \phi$ is similar to $T$. Typical registration approaches (Sotiras et al., 2013) are formulated as an optimization problem that aims to find the most optimized $\phi$ to minimize the energy:

$$\phi = \underset{\phi}{\operatorname{argmin}}\, M(T, S \circ \phi) + R(\phi). \tag{1}$$

The first term $M(T, S \circ \phi)$ quantifies the distance between the template image and the warped subject image. The second term $R(\phi)$ regularizes $\phi$ so that it is well-behaved.

In this paper, we present a novel *hierarchical dual-supervised FCN* for brain deformable registration (see **Fig. 1** for overview). Our implementation is based on overlapping $64 \times 64 \times 64$ image patches. The output is $24 \times 24 \times 24$ patch of displacement vectors, because the deformable prediction is highly related to the local information of the image and also we can only estimate the deformation field in the center region. For the end-to-end prediction of the deformation field from an image, we utilize a U-Net (Ronneberger et al., 2015) based regression model. In particular, we propose four strategies to improve registration:

1) **Hierarchical dual-supervision.** In addition to deformation fields, we use the difference between images as additional information for supervising the training. We also employ hierarchical loss layers in the upsampling path of U-Net, giving more constraint in the frontal layers for easier convergence.

2) **Gap filling.** To improve prediction accuracy, additional convolutional layers are further inserted between the u-type ends to connect low-level and high-level features.

3) **Multi-channel inputs**. In addition to image intensity, difference map and gradient maps are also used as inputs to the network.

4) **Data augmentation.** To overcome over-fitting, training data are augmented by warping the images with different degrees and by iteratively including predicted images in training.

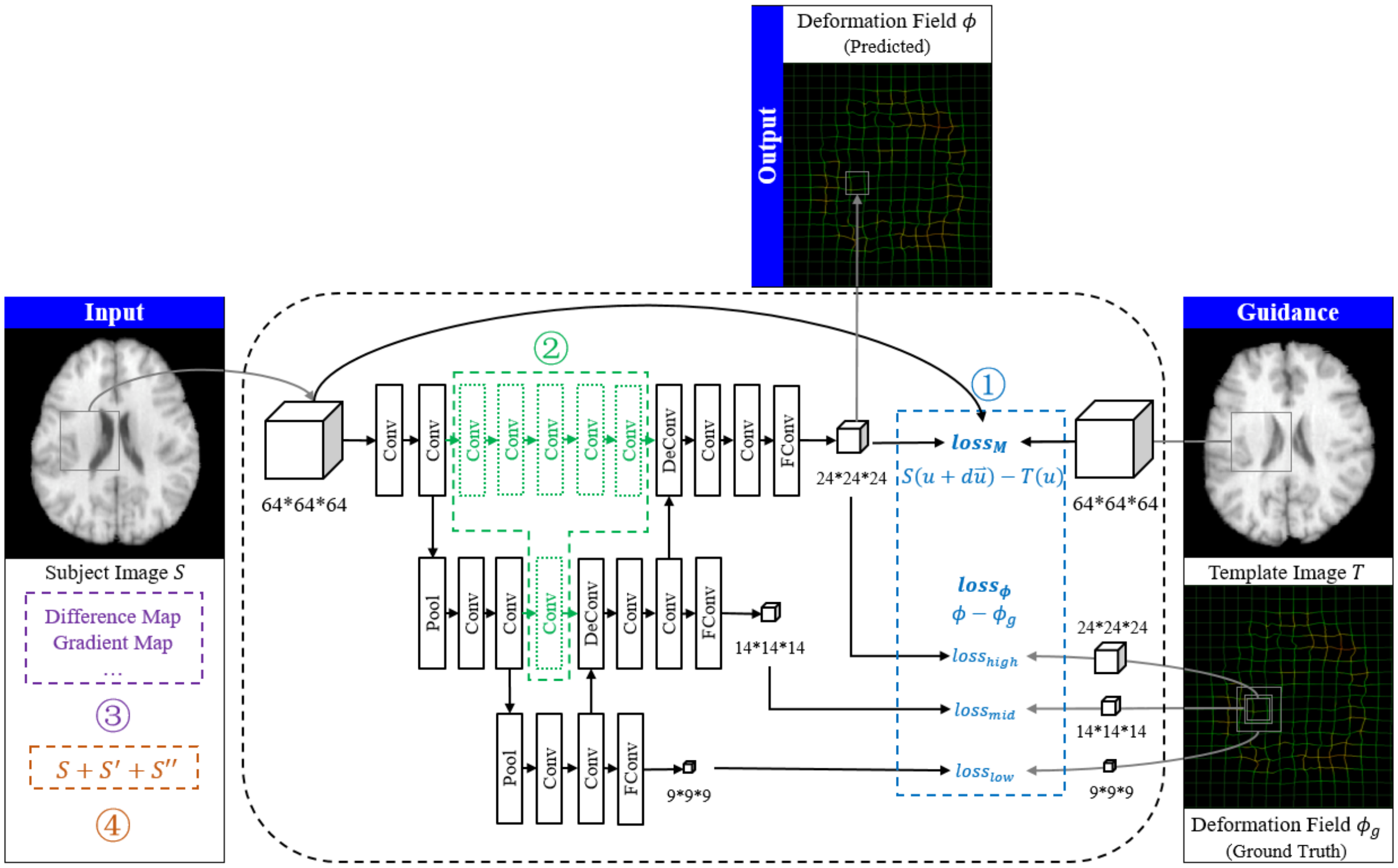

**Fig. 1.** Overview of our proposed method.

### 3.2 Hierarchical Dual-Supervised FCN

#### 3.2.1 Hierarchical Dual-Supervision

a) Dual-supervision

In our dual-guidance strategy, the loss function consists of two parts: 1) $loss_\phi$ - the difference between the predicted deformation field and the existing (training) ground-truth deformation field; 2) $loss_M$ - the difference between the template and the warped subject image based on the deformation currently estimated via the network.

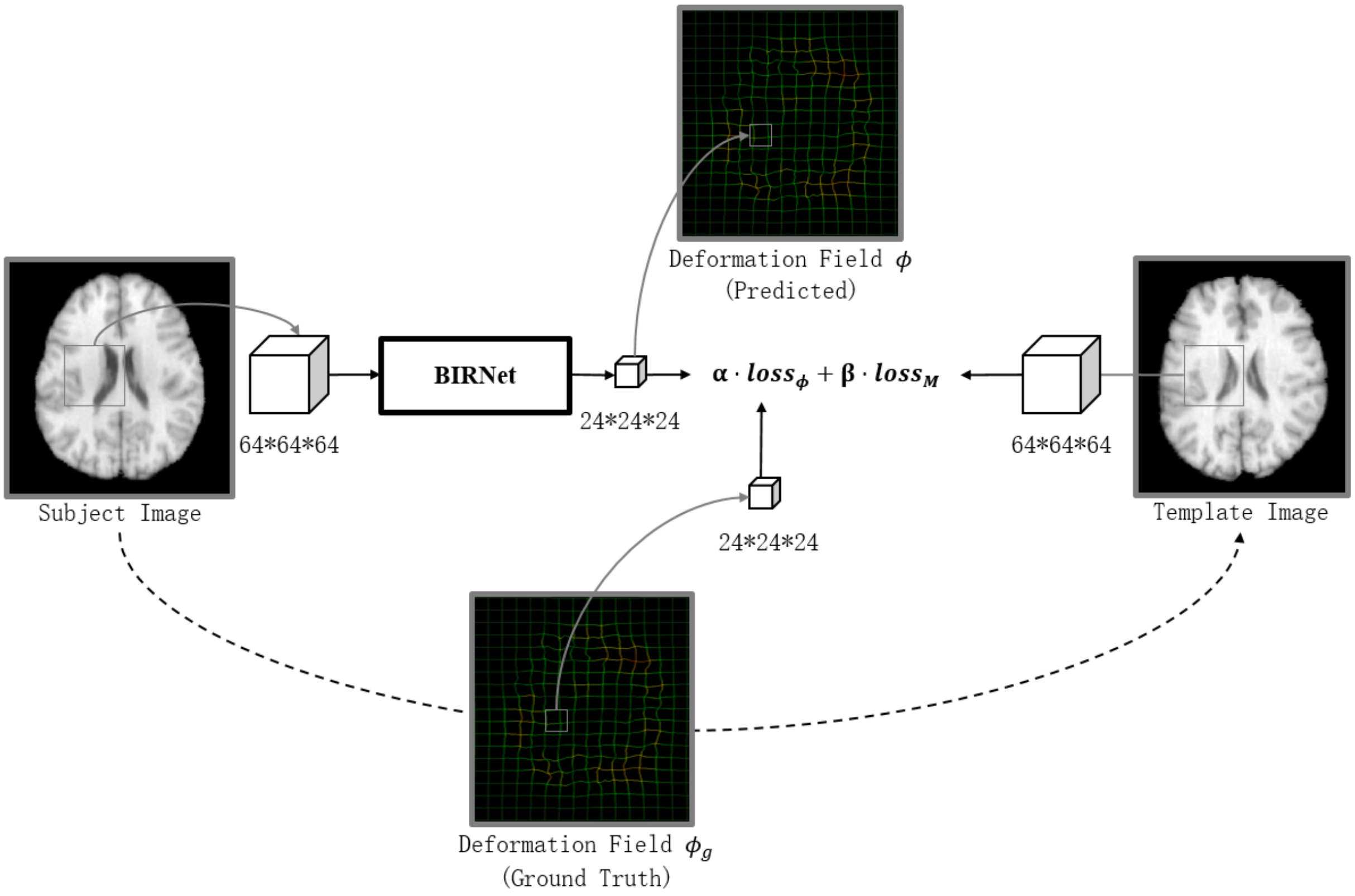

**Fig. 2.** Training strategy with loss function $\alpha \cdot loss_\phi + \beta \cdot loss_M$.

Here, $loss_\phi$ is a the Euclidean distance as defined in (Rohé et al., 2017; Sokooti et al., 2017), which assumes that the ground-truth deformation fields are already achieved. As shown in Fig. 2, for the template image $T$ and subject image $S$, the ground-truth deformation field $\phi_g$ is used to guide the training of the deep learning model with loss function:

$$loss_\phi = \frac{1}{N}\left\|\phi - \phi_g\right\|_2^2 \tag{2}$$

where $\phi$ is the predicted deformation field and $N$ is the number of voxels. Note that the performance of trained model is limited by the ground-truth deformation fields, which are obtained using traditional registration methods before training (Avants et al., 2008; Vercauteren et al., 2009). To improve accuracy, we include the following loss function:

$$loss_M = \frac{1}{N}\sum_u \left\|S\left(u + \phi(u)\right) - T(u)\right\|_2^2 \tag{3}$$

where $u$ represents the voxel coordinate $[x, y, z]$ in the template space and $\phi(u) = \left[d_x, d_y, d_z\right]$ is the displacement of $u$. The final loss function is

$$loss = \alpha \cdot loss_\phi + \beta \cdot loss_M \ , \ \alpha + \beta = 1 \tag{4}$$

where α and $\beta$ are the two coefficients satisfying $\alpha \geq 0$, $\beta \geq 0$, and $\alpha + \beta = 1$. They are dynamically varied during the training process, i.e., taking a larger $\alpha$-value in the initial training stage to accelerate convergence and achieve smoother deformation fields and then taking a larger $\beta$-value in the fine-tuning stage to refine the registration. In actual implementation, we first normalize the range of two loss functions to $[0,1]$. Then, we set $\alpha = 0.8$ and $\beta = 0.2$ in the initial training stage (i.e., the first 5 epochs), and $\alpha = 0.5$ and $\beta = 0.5$ in the later training stage (i.e., the last 5 epochs).

Specifically, the loss function works by giving the gradient value backward to the front layers. Let $[d_x, d_y, d_z]$ consist of a vector of displacements. Then, the gradient in $x$-direction can be represented by the following equation:

$$\frac{\partial loss_M}{\partial dx} = \frac{M(dx+\Delta dx)-M(dx)}{\Delta dx} = \begin{array}{c} |S(x + dx + 1, y + dy, z + dz) - T(x, y, z)| \\ -|S(x + dx, y + dy, z + dz) - T(x, y, z)| \end{array} \quad (5)$$

where we calculate an error in recent vector first, and then plus one in the recent direction $x$ and calculate the varying error. Finally, the difference between them will be the gradient. The gradient of $d_y$ and $d_z$ could be calculated in the same way.

In summary, using dual-guidance can effectively combine the advantages of both loss functions: 1) the rough guidance provided by $loss_\phi$ makes the convergence easily and fast; and 2) the image difference guidance provided by $loss_M$ further refines the registration results, which can address the issue of inaccurate ground truth.

b) Hierarchical supervision

In the conventional U-Net, the loss is calculated only in the final layer, resulting in suboptimal parameters in the frontal convolution layers (Schmidhuber, 2015). In this way, the parameters of the first half of the convolution layers are not updated as much as the latter half. This *not only* causes slow convergence, *but also* over-fitting. Therefore, we add a loss function in each of the layer to directly supervise the training of the first (frontal) half of the network.

As we use filters with size $3 \times 3 \times 3$, each convolutional layer without padding reduces the patch size isotropically by one voxel. Also, each pooling layer will further downsample the patch. As a result, for an input patch size of $64 \times 64 \times 64$, we extract $24 \times 24 \times 24$ patch $\phi_g^{high}$ for high resolution, $14 \times 14 \times 14$ patch $\phi_g^{mid}$ for middle resolution, and $9 \times 9 \times 9$ patch $\phi_g^{low}$ for low resolution. The translations from $\phi_g$ to $\phi_g^{high}$, $\phi_g^{mid}$, $\phi_g^{low}$ are:

$$\begin{aligned} \phi_g^{high}(i,j,k) &= \phi_g(i+20, j+20, k+20) & i,j,k \in [0,23] \\ \phi_g^{mid}(i,j,k) &= \phi_g(i \times 2+18, j \times 2+18, k \times 2+18)/2 & i,j,k \in [0,13] \\ \phi_g^{low}(i,j,k) &= \phi_g(i \times 4+14, j \times 4+14, k \times 4+14)/4 & i,j,k \in [0,8] \end{aligned} \tag{6}$$

where $i, j, k$ are the coordinates of the points in the patch. Then, we can easily calculate the respective loss function for each level as:

$$\begin{aligned} loss_\phi^{high} &= \frac{1}{24 \times 24 \times 24} \left\| \phi^{high} - \phi_g^{high} \right\|_2^2 \\ loss_\phi^{mid} &= \frac{1}{14 \times 14 \times 14} \left\| \phi^{mid} - \phi_g^{mid} \right\|_2^2 \\ loss_\phi^{low} &= \frac{1}{9 \times 9 \times 9} \left\| \phi^{low} - \phi_g^{low} \right\|_2^2 \end{aligned} \tag{7}$$

where $\phi^{high}$, $\phi^{mid}$, $\phi^{low}$ are predicted by the learning model directly, which are in the same size with $\phi_g^{high}$, $\phi_g^{mid}$, $\phi_g^{low}$. Finally, the total loss function $loss_\phi$ is:

$$loss_\phi = loss_\phi^{high} + loss_\phi^{mid} + loss_\phi^{low} \tag{8}$$

### 3.2.2 Gap filling

The black network in **Fig. 3** is the basic network architecture of U-Net. It consists of a contracting path (left side) and an expansive path (right side). The contracting path follows the typical architecture of a convolutional network. It consists of repeated applications of two $3 \times 3 \times 3$ convolutions (i.e., convolutions that may be followed by ReLU (He et al., 2015), and batch normalization layers (Ioffe and Szegedy, 2015)), and a $2 \times 2 \times 2$ max pooling operation with stride size of 2 for downsampling. At each downsampling step, we double the number of feature channels. Every step in the expansive path consists of a $2 \times 2 \times 2$ deconvolution to upsample the feature map and halve the number of feature channels, and also two $3 \times 3 \times 3$ convolutions. At the final layer, a $1 \times 1 \times 1$ convolution is used to map each 64-component feature vector to the desired number of classes. To recover detail lost due to downsampling, we concatenate the correspondingly cropped feature map from the contracting path.

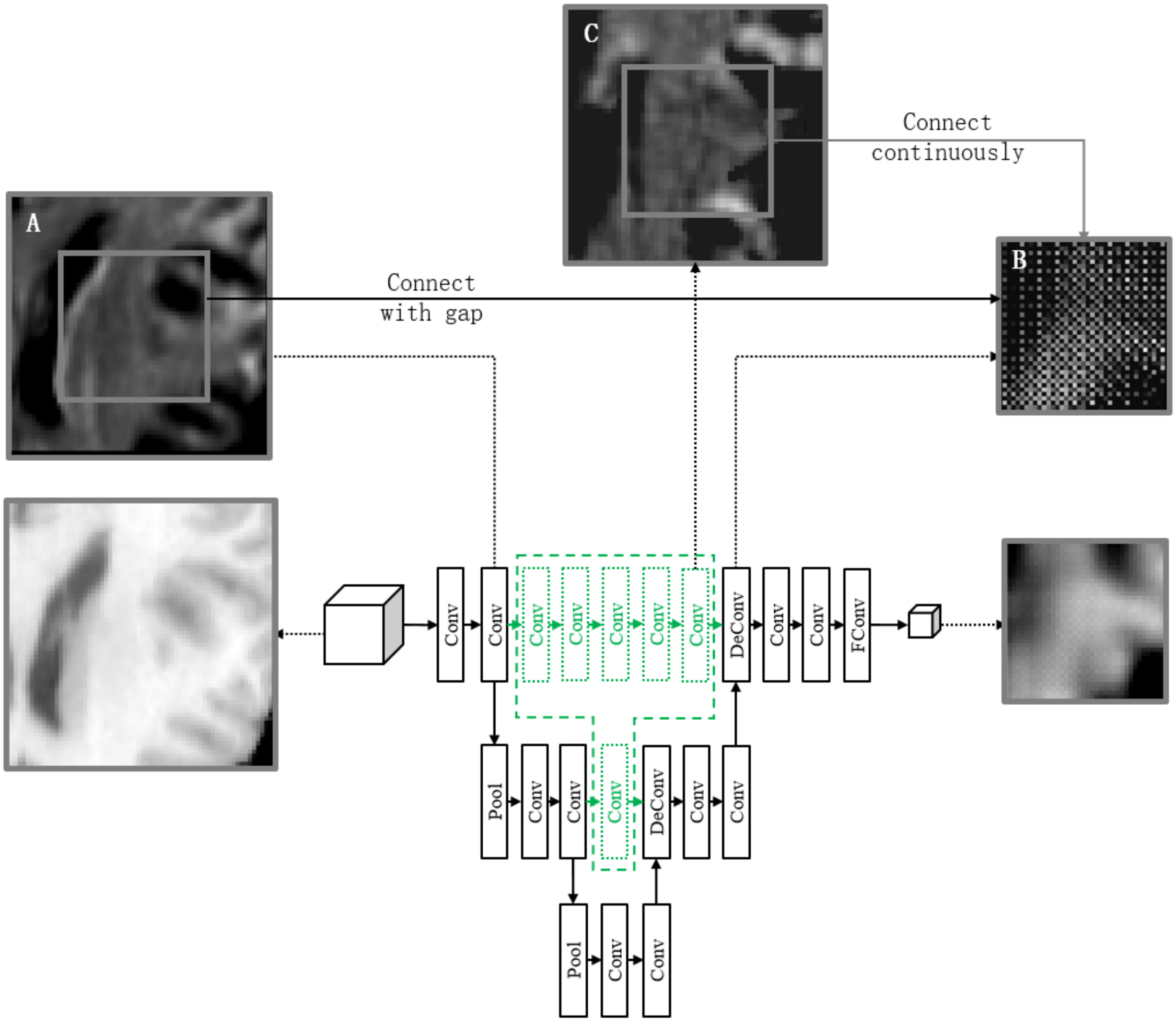

**Fig. 3.** The feature image samples of the typical layers in the network. The output deformation field is shown by the feature image of the displacement value in *x*-axial. The noise-like pattern in image B is caused by the upsampling process in deconvolution layer.

As shown in **Fig. 3**, two feature maps A and B are significantly dissimilar. Feature map A resembles the original image, whereas feature map B resembles the deformation field. Obviously, there is a huge gap between the feature maps A and B, which are usually concatenated together by the conventional U-Net. This gap makes the network less efficient in both training and testing stages.

To address this issue, we propose to include additional convolution layers between the same level of the contracting phase and the expansion phase (as shown by the green network in Fig. 3) to synchronize the convolution path of the feature maps. The parameters of the added convolution layers are equal to the convolution layers in the lower resolution. In this way, the feature map C after gap filling will be more similar to the feature map B, improving both registration accuracy and training speed.

### 3.2.3 Multi-Channel Inputs

Image feature maps, such as difference and gradient maps, can also be used to improve registration accuracy. Although these can be learned in the deep learning network, computation resource is needed to learn them. **Fig. 4** shows the multi-channel inputs, including original image, difference map, and gradient map. The different map is computed as the intensity difference between the subject and template images. The gradient map provides boundary information to help structural alignment. Moreover, the gradient maps are used to calculate the image similarity in equation (3) in addition to the original image. Then, this constraint ensures that the boundary regions could be aligned more accurately than only matching the image similarity in the original intensity image.

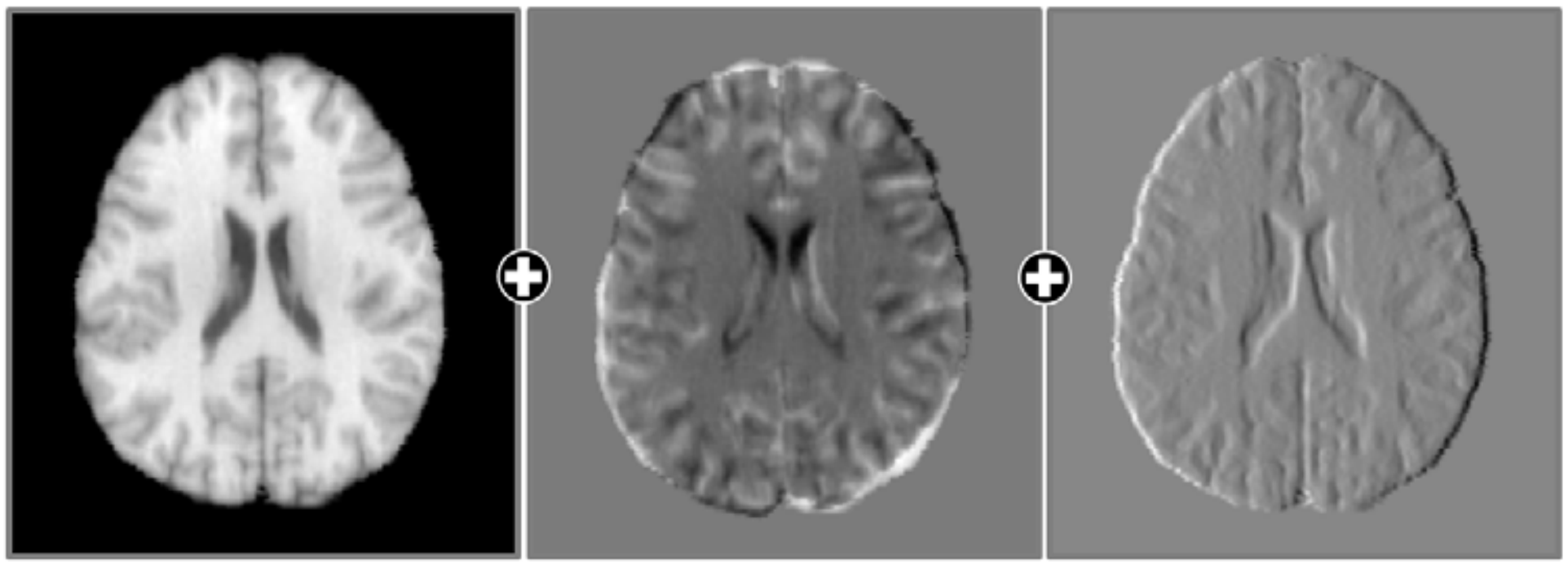

**Fig. 4.** The concatenated original image, difference map, and gradient map.

### 3.2.4 Implementation

**Fig. 5** shows the architecture of the proposed 3D image registration network, which is based on the $64 \times 64 \times 64$ input patches of image appearance and the $24 \times 24 \times 24$ output patches of deformation. The network is implemented using 3D Caffe (Jia et al., 2014) and optimized using Adam (Kingma and Ba, 2014). We set the learning rate to 1e-3 for the initial training stage and 1e-8 for the fine-tuning training stage. The network takes one 3D patch from the subject image as the input, and outputs one 3D deformation field patch, which consists of three independent patches for the *x*, *y* and *z* dimensions, respectively (note that only one branch is shown in **Fig. 5**).

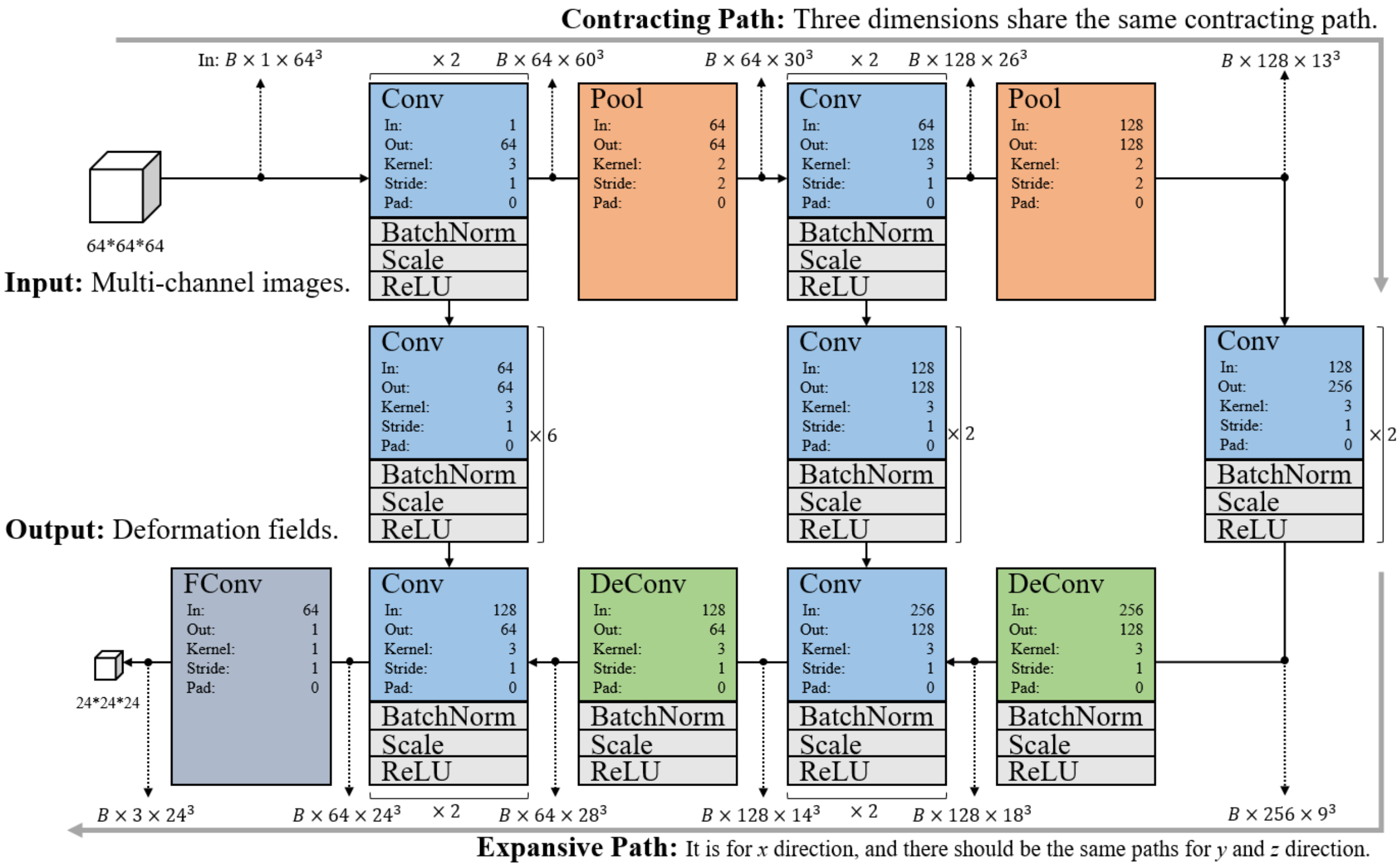

**Fig. 5.** The architecture of image registration network. Conv: 3D convolution layer. Pool: 3D pooling layer. DeConv: Deconvolution layer. BatchNorm: Batch normalization layer. Scale: Scale layer. ReLU: ReLU layer. In: The number of input channels. Out: The number of output channels. Kernel: The kernel size of the 3D filter in each dimension. Stride: Stride of the 3D filter. Pad: Zero-padding. *B*: Batch size.

### 3.3 Data Augmentation

We evaluate our method on 3D brain MR images. We use LONI LPBA40 (Shattuck et al., 2008) dataset (image size: $220 \times 220 \times 184$) for training, where we choose one image as the template, 30 images as the training images, and the remaining 9 images as the validation images. The ground-truth deformation fields for all training and validation data are generated using both Diffeomorphic Demons (Vercauteren et al., 2009) and SyN (Avants et al., 2008).

We augment the dataset because it is too small for effective training. This is done by warping each subject image with 20%, 40%, 60%, 80%, and 100% of the ground truth deformation, meanwhile the respective deformation field is the target of prediction. An example is shown in **Fig. 6**. It works because the deep learning model does not iteratively calculate the deformation field, and thus the intermediate results with different degrees of deformation are effective samples for deep network training. This significantly expands the size of training dataset by 6 folds, and will allow coarse and fine deformations to participate in the training.

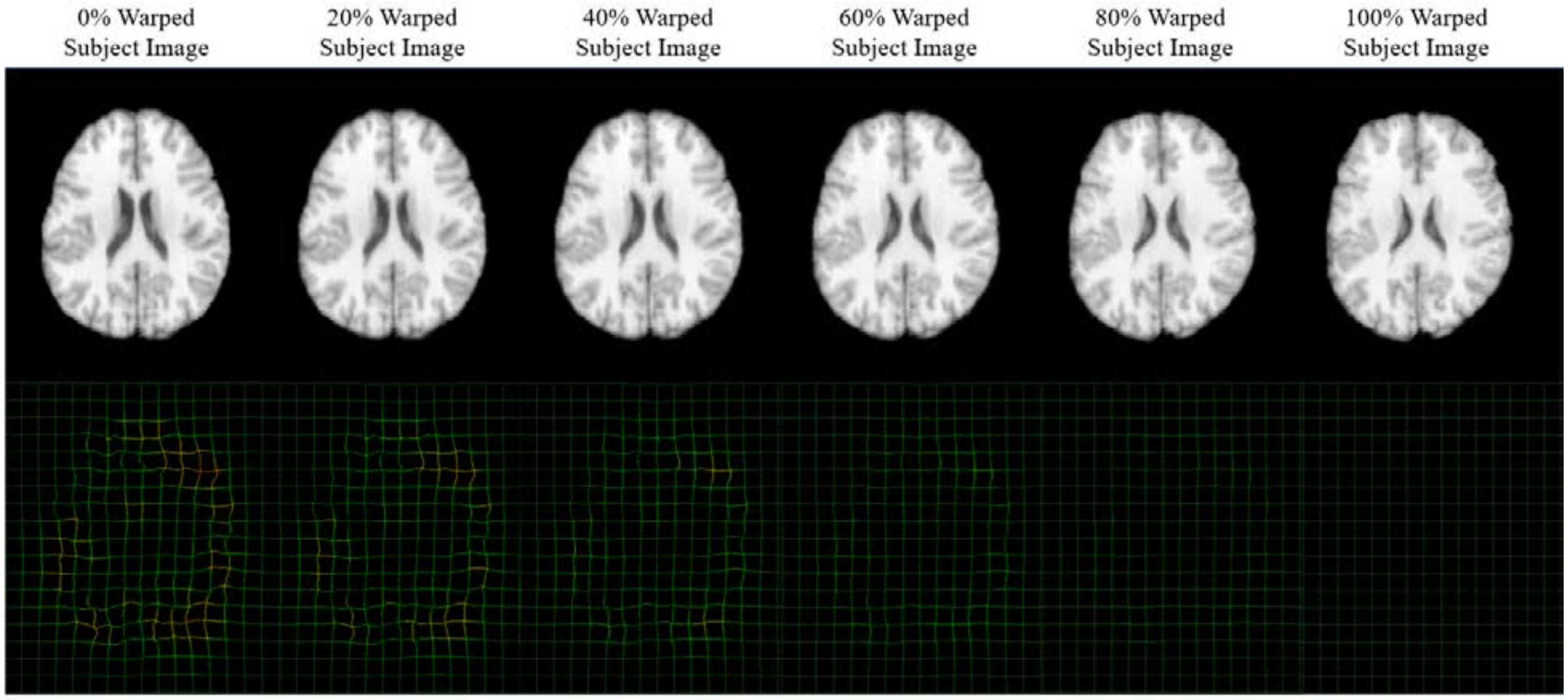

**Fig. 6.** Expanded training data constructed by warping the subject image with varying degrees.

## 4. Experiments

To evaluate the performance of our proposed method, the comparison with several state-of-the-art deformable registration algorithms is shown in this section. We train BIRNet using LPBA40 and then apply it to four different datasets, including IBSR18 (Klein et al., 2009), CUMC12 (Klein et al., 2009), MGH10 (Klein et al., 2009), and IXI30 (Serag et al., 2012). We select 3 state-of-the-art registration methods (Klein et al., 2009), i.e., Diffeomorphic Demons (Vercauteren et al., 2009), FNIRT (Andersson et al., 2007), and SyN (Avants et al., 2008), for comparison.

### 3.1 Evaluation based on LPBA40

We test the performance of BIRNet on LPBA40 dataset. For each of the 180 training images, we extract 300 patches of size $64 \times 64 \times 64$, giving us a total of 54000 training patches. Three models are considered: 1) The original **U-Net**, 2) BIRNet without similarity guidance, denoted as **BIRNet_WOS**, and 3) **BIRNet** using dual-guidance (our proposed method). **Fig. 7** shows the loss curves of $loss_{\phi}$ the $loss_{M}$ for both training and validation. We can see from **Fig.7 (a)** that the performance of U-Net saturates as training progresses. BIRNet_WOS improves the performance, but the best performance is given by BIRNet, which further considers image similarity/difference. Meanwhile, the BIRNet model has reached a better accuracy even than the ground truth in **Fig.7**

**(b)**. These results demonstrate that the image similarity/difference loss can provide useful guidance to further refine the training model, even the ground-truth deformation fields cannot be quite accurate. **Fig. 8** shows an example of the registration outcome, confirming that BIRNet produces results that are most similar to the template especially in the yellow squared regions.

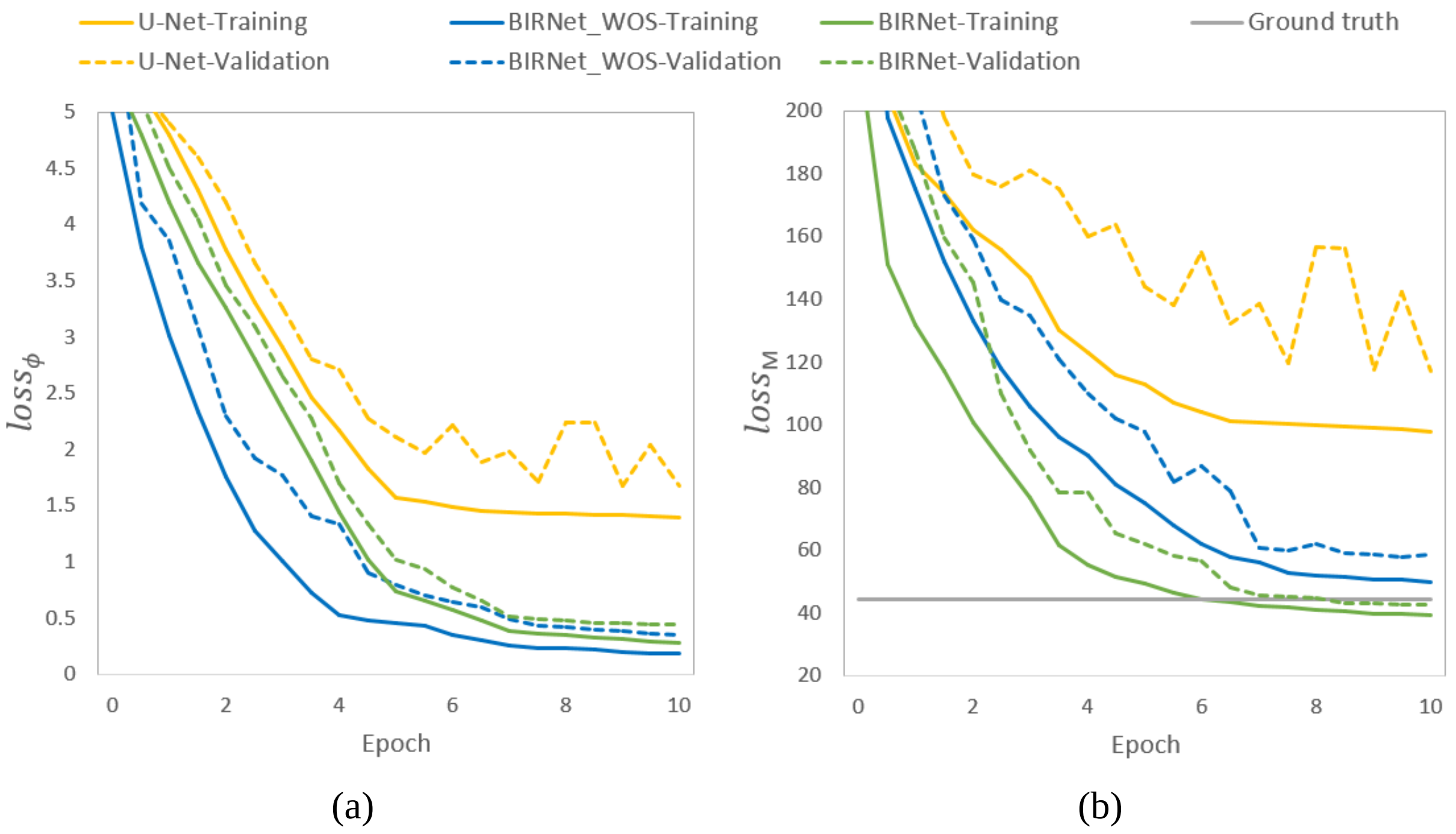

**Fig. 7.** The training and validation curves for (a) $loss_\phi$ and (b) $loss_M$. The value of $loss_\phi$is shown in mean square error of *displacement*, meanwhile the value of $loss_M$ is shown in mean square error of *intensity*.

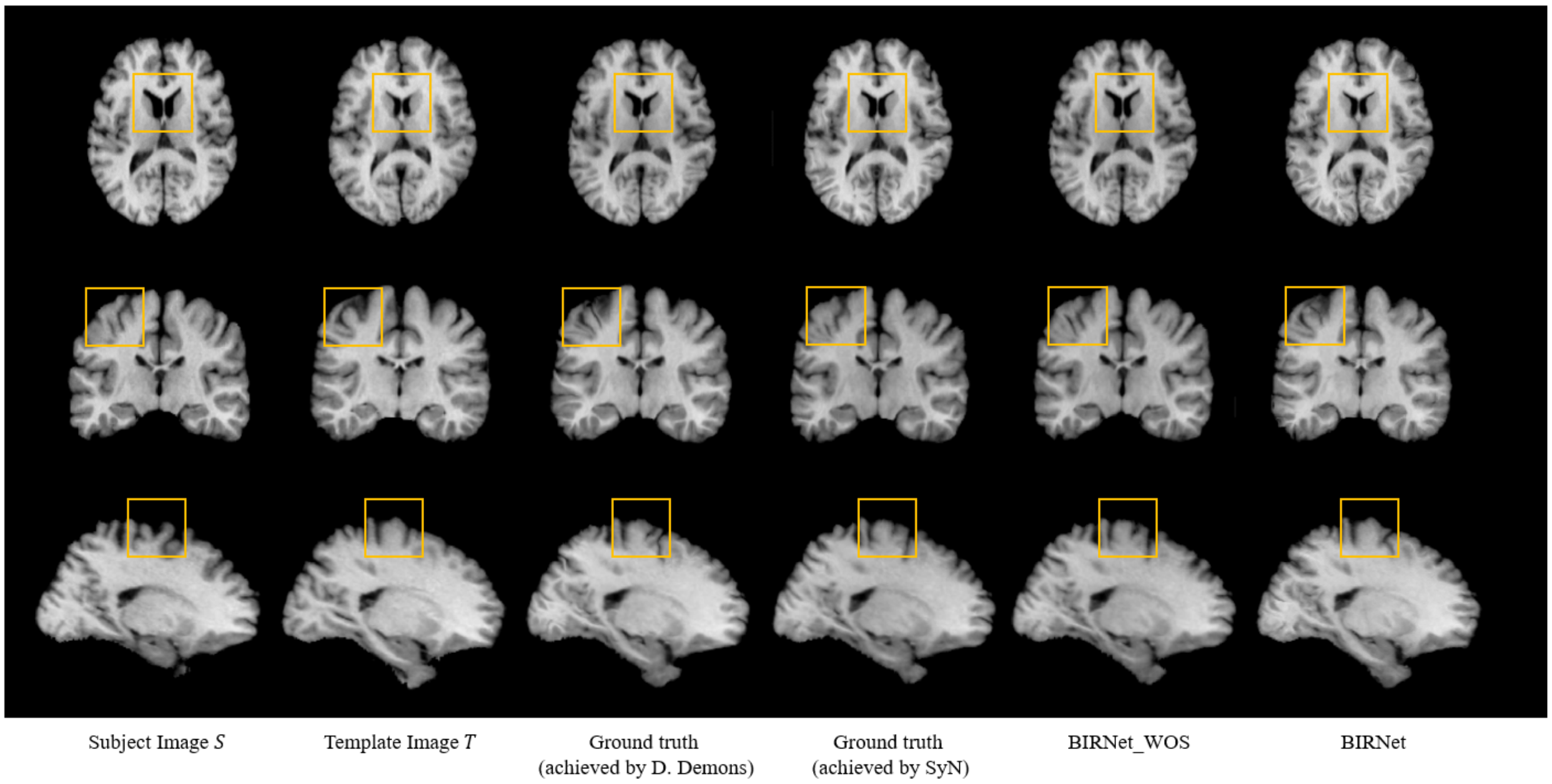

**Fig. 8.** An example of the registration outcomes. Improvements are marked by yellow boxes.

**Fig. 9** shows the Dice similarity coefficient (DSC) of 54 brain ROIs (with the ROI names give in Table 1, obtained from (Shattuck et al., 2008)). We observe that BIRNet yields better performance for 35 out of 54 ROIs and the comparable performance for the other 19 ROIs with Diffeomorphic Demons and SyN. BIRNet_WOS shows accuracy that is a little worse than Diffeomorphic Demons and SyN, which shows that the dual-guidance is effective in boosting the performance.

**Table 1.** The names of the ROIs in LONI LPBA40 dataset.

| ID | Fullname | ID | Fullname | ID | Fullname |
|---|---|---|---|---|---|
| 1 | L superior frontal gyrus | 19 | L supramarginal gyrus | 37 | L inferior temporal gyrus |
| 2 | R superior frontal gyrus | 20 | R supramarginal gyrus | 38 | R inferior temporal gyrus |
| 3 | L middle frontal gyrus | 21 | L angular gyrus | 39 | L parahippocampal gyrus |
| 4 | R middle frontal gyrus | 22 | R angular gyrus | 40 | R parahippocampal gyrus |
| 5 | L inferior frontal gyrus | 23 | L precuneus | 41 | L lingual gyrus |
| 6 | R inferior frontal gyrus | 24 | R precuneus | 42 | R lingual gyrus |
| 7 | L precentral gyrus | 25 | L superior occipital gyrus | 43 | L fusiform gyrus |
| 8 | R precentral gyrus | 26 | R superior occipital gyrus | 44 | R fusiform gyrus |
| 9 | L middle orbitofrontal gyrus | 27 | L middle occipital gyrus | 45 | L insular cortex |
| 10 | R middle orbitofrontal gyrus | 28 | R middle occipital gyrus | 46 | R insular cortex |
| 11 | L lateral orbitofrontal gyrus | 29 | L inferior occipital gyrus | 47 | L cingulate gyrus |
| 12 | R lateral orbitofrontal gyrus | 30 | R inferior occipital gyrus | 48 | R cingulate gyrus |
| 13 | L gyrus rectus | 31 | L cuneus | 49 | L caudate |
| 14 | R gyrus rectus | 32 | R cuneus | 50 | R caudate |
| 15 | L postcentral gyrus | 33 | L superior temporal gyrus | 51 | L putamen |
| 16 | R postcentral gyrus | 34 | R superior temporal gyrus | 52 | R putamen |
| 17 | L superior parietal gyrus | 35 | L middle temporal gyrus | 53 | L hippocampus |
| 18 | R superior parietal gyrus | 36 | R middle temporal gyrus | 54 | R hippocampus |

**Fig. 10** shows the DSC results for 9 validation subject images from LPBA40. BIRNet_WOS results in a slight performance drop compared with Diffeomorphic Demons and SyN, but only by a small extent (i.e., less than 1.5% in average). BIRNet again achieves the best performance with higher DSC values on 29 out of 54 ROIs and very similar values on the other 25 ROIs compared with Diffeomorphic Demons and SyN. The average DSC of Diffeomorphic Demons, SyN, BIRNet_WOS and BIRNet are 67.3%, 68.1%, 66.7% and 69.2, respectively. These results verify the generalizability of BIRNet.

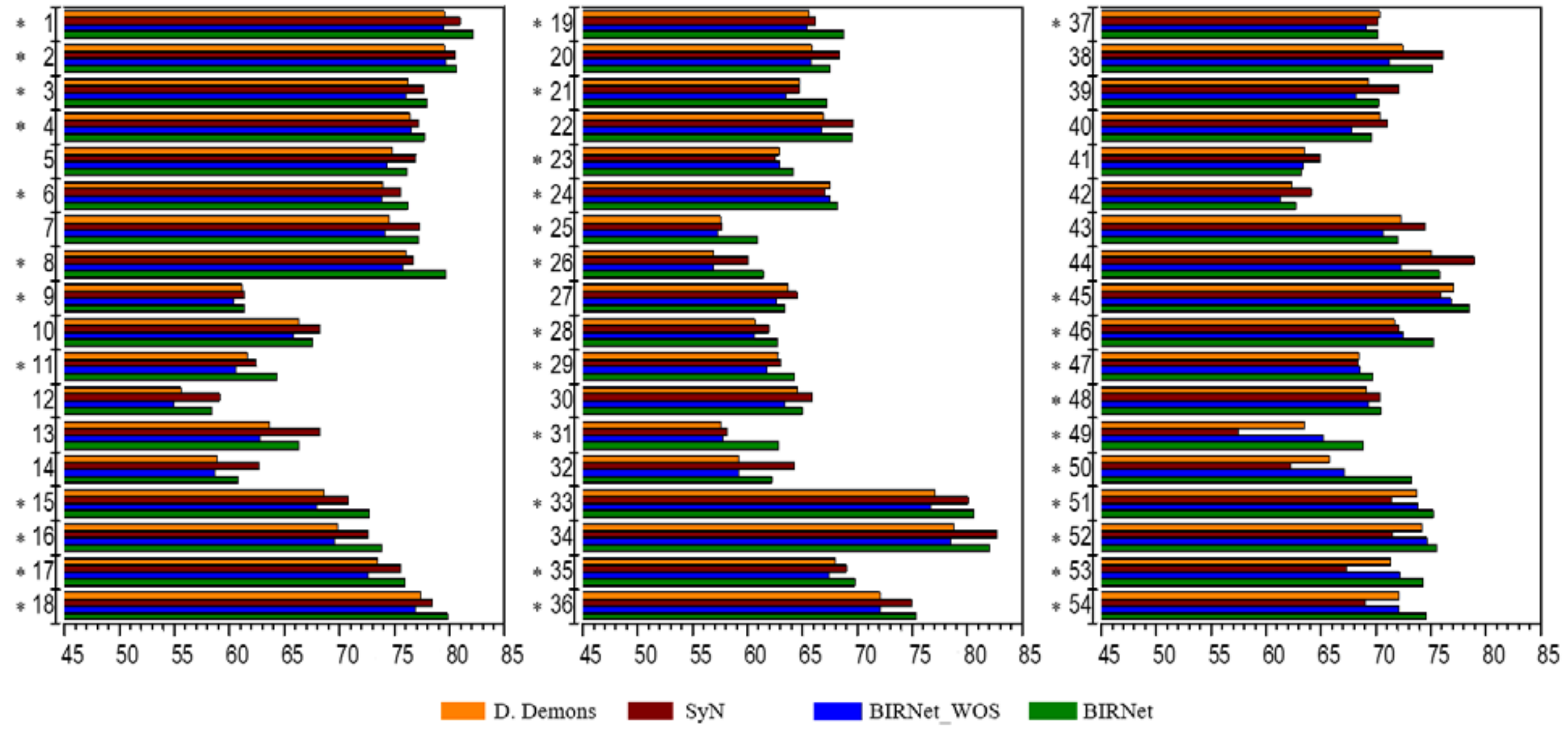

**Fig. 9.** The mean DSCs of 54 ROIs based on 30 training subjects from the LPBA40 dataset, after deformable registration by Diffeomorphic Demons, SyN, BIRNet_WOS and BIRNet. "*" marks the cases where BIRNet achieves the highest DSC value among all the four methods.

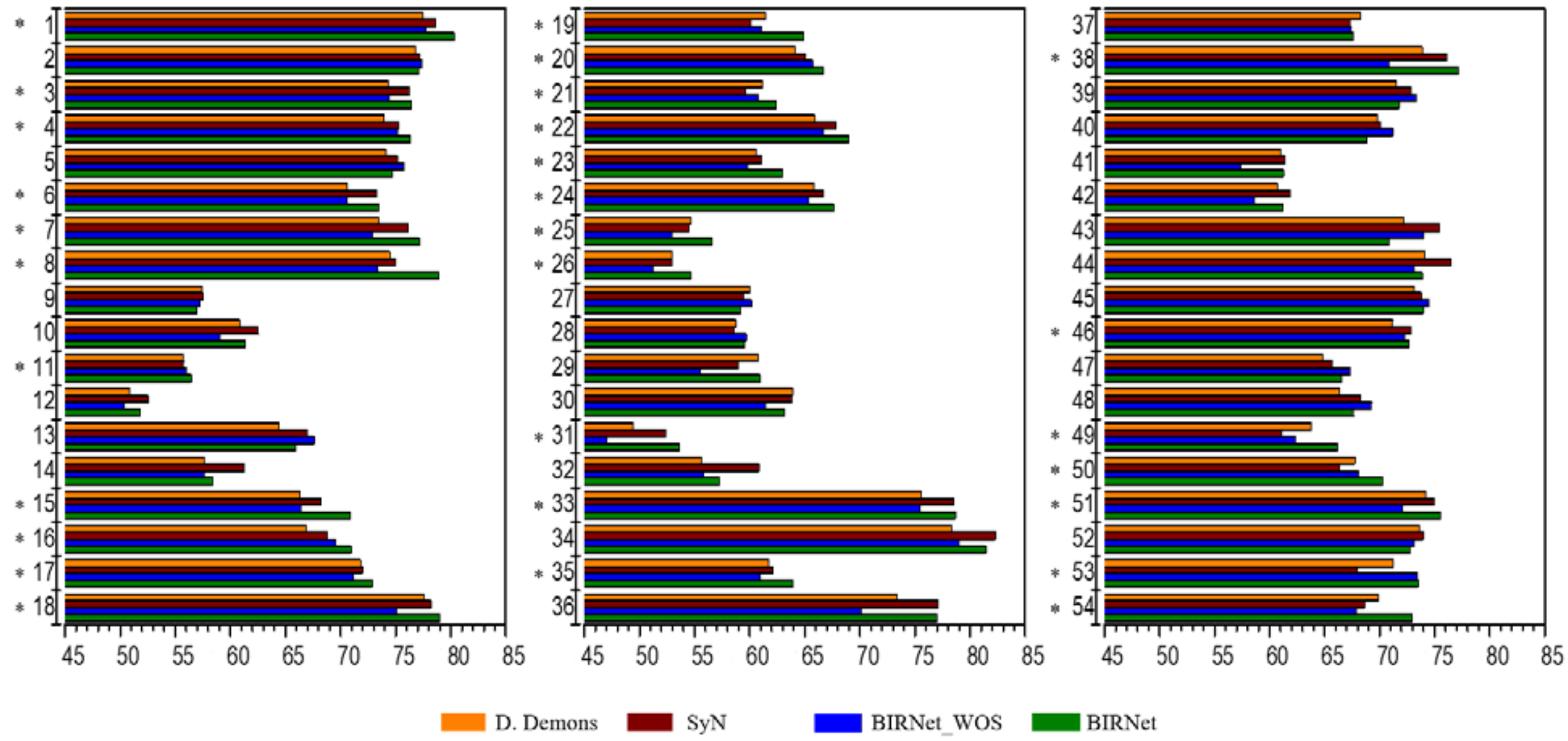

**Fig. 10.** The mean DSC of 54 ROIs based on 9 validation subjects from LPBA40 dataset, after deformable registration by Diffeomorphic Demons, SyN, BIRNet_WOS and BIRNet. "*" marks the cases where BIRNet achieves the highest DSC value among all the four methods.

### 3.2 Evaluation based on IBSR18, CUMC12, MGH10, IXI30

To further evaluate the accuracy and generalizability of BIRNet, we further test it on a total of 70 brain images from four different datasets, i.e., IBSR18 (Klein et al., 2009), CUMC12 (Klein et al., 2009), MGH10 (Klein et al., 2009), and IXI30 (Serag et al., 2012), by directly applying the model trained using the LPBA40 dataset without any additional parameter tuning. The results for one subject of the IBSR18 dataset is shown in **Fig.11** for Diffeomorphoc Demons, SyN, FNIRT (Andersson et al., 2007) and BIRNet. Note that, the results shown for SyN and FNIRT are based on their optimal parameters determined individually for each image. **Table 2** provides the DSCs for of Gray Matter (GM) and White Matter (WM) based on GM and WM labels provided in these four datasets. The performance of BIRNet is comparable to the ***fine-tuned*** SyN and FNIRT (particularly to each of these four datasets), but without the need for parameter tuning. This verifies the generalizability of BIRNet.

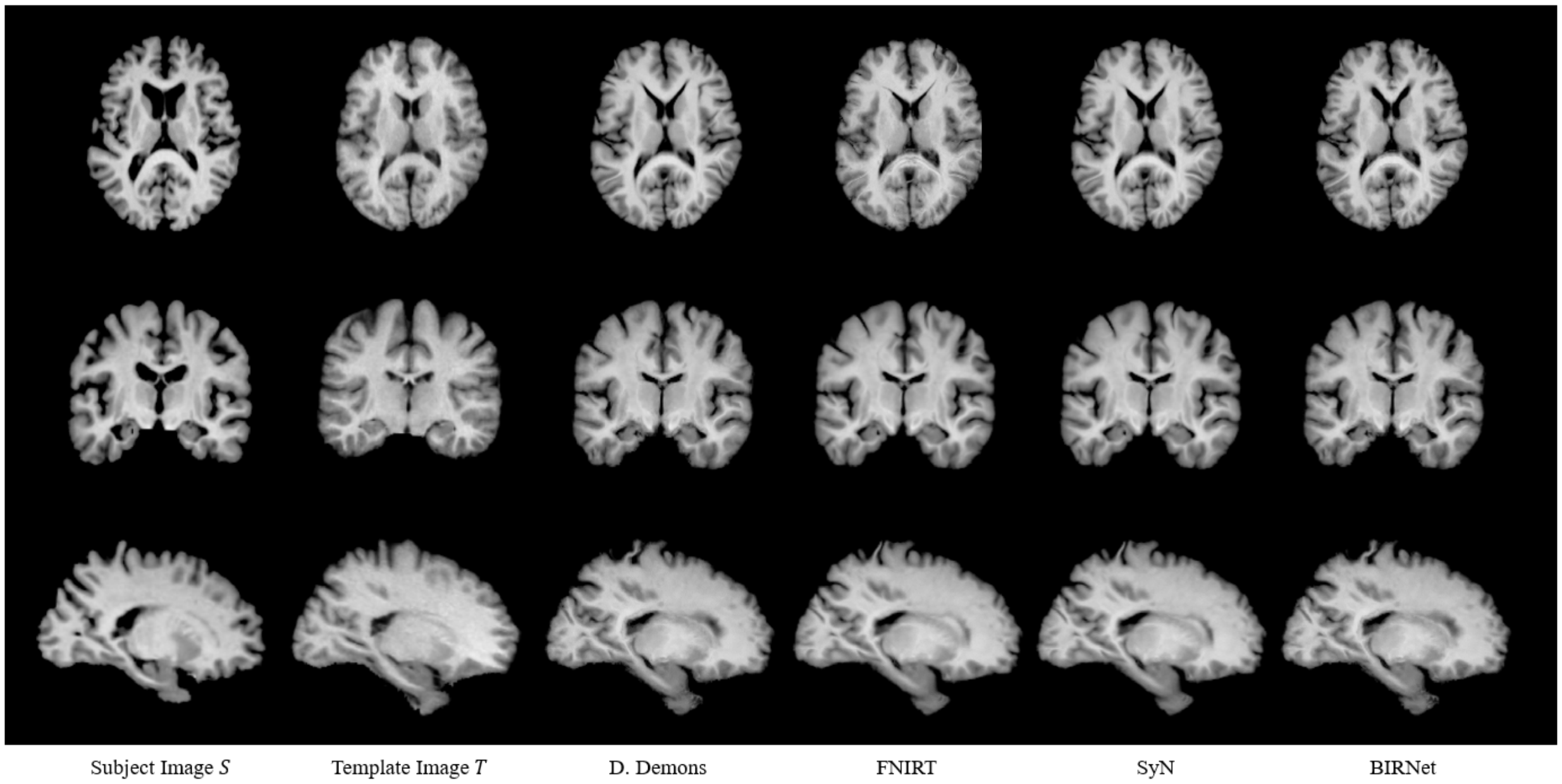

**Fig. 11.** Example testing case in IBSR18 dataset.

**Table 2.** Results for IBSR18, CUMC12, MGH10, IXI30 in term of DSC (%). Results for both default and tuned parameters are shown for FNIRT and SyN.

| Dataset | Brain Tissue | **D.Demons** | **FNIRT** | **FNIRT (default)** | **SyN** | **SyN (default)** | **BIRNet** |
|---|---|---|---|---|---|---|---|
| IBSR18 | GM | 73.7±2.4 | **74.3±1.8** | 73.1±2.3 | 73.9±2.2 | 72.9±2.8 | 74.1±2.3 |
| | WM | 75.8±1.5 | 76.5±2.0 | 75.1±1.9 | **77.6±1.7** | 75.2±2.3 | 77.0±2.0 |
| CUMC12 | GM | 74.6±2.2 | 74.4±2.4 | 73.4±3.1 | **75.1±1.8** | 73.2±3.4 | 73.8±2.7 |
| | WM | 75.5±2.0 | 76.3±1.5 | 74.9±2.0 | **76.7±1.2** | 74.3±2.2 | 76.5±1.4 |
| MGH10 | GM | 73.1±3.4 | **74.1±2.8** | 73.1±3.1 | 73.6±2.3 | 72.8±4.1 | 73.4±2.9 |
| | WM | 78.3±1.6 | 78.8±2.1 | 77.9±2.5 | 79.1±1.9 | 77.7±2.8 | **79.6±1.7** |
| IXI30 | GM | 72.4±2.5 | 74.4±2.4 | 72.7±2.5 | **75.2±1.6** | 71.6±2.9 | 73.7±2.1 |
| | WM | 76.9±1.9 | 78.1±2.0 | 77.4±2.7 | **78.3±1.7** | 76.5±3.0 | 77.8±1.9 |

### 3.4 Computation costs

BIRNet is implemented based on Caffe (Jia et al., 2014) on a single Nvidia TitanX (Pascal) GPU. For fair comparison, we compare its speed with CPU and GPU implementations of other comparison methods. (Note that there is no GPU implementation for FNIRT (Fluck et al., 2011)). **Fig. 12** shows the computation costs for a typical 3D brain image (220×220×184) of seven different deformable registration algorithms: Diffeomorphic Demons - CPU (Vercauteren et al., 2009), Diffeomorphic Demons - GPU (Muyan-Ozcelik et al., 2008), SyN - CPU (Avants et al., 2008), SyN - GPU (Luo et al., 2015), FNIRT - CPU (Andersson et al., 2007), and BIRNet. It is clear that BIRNet, which does not require any iterative optimization, requires the least amount of time.

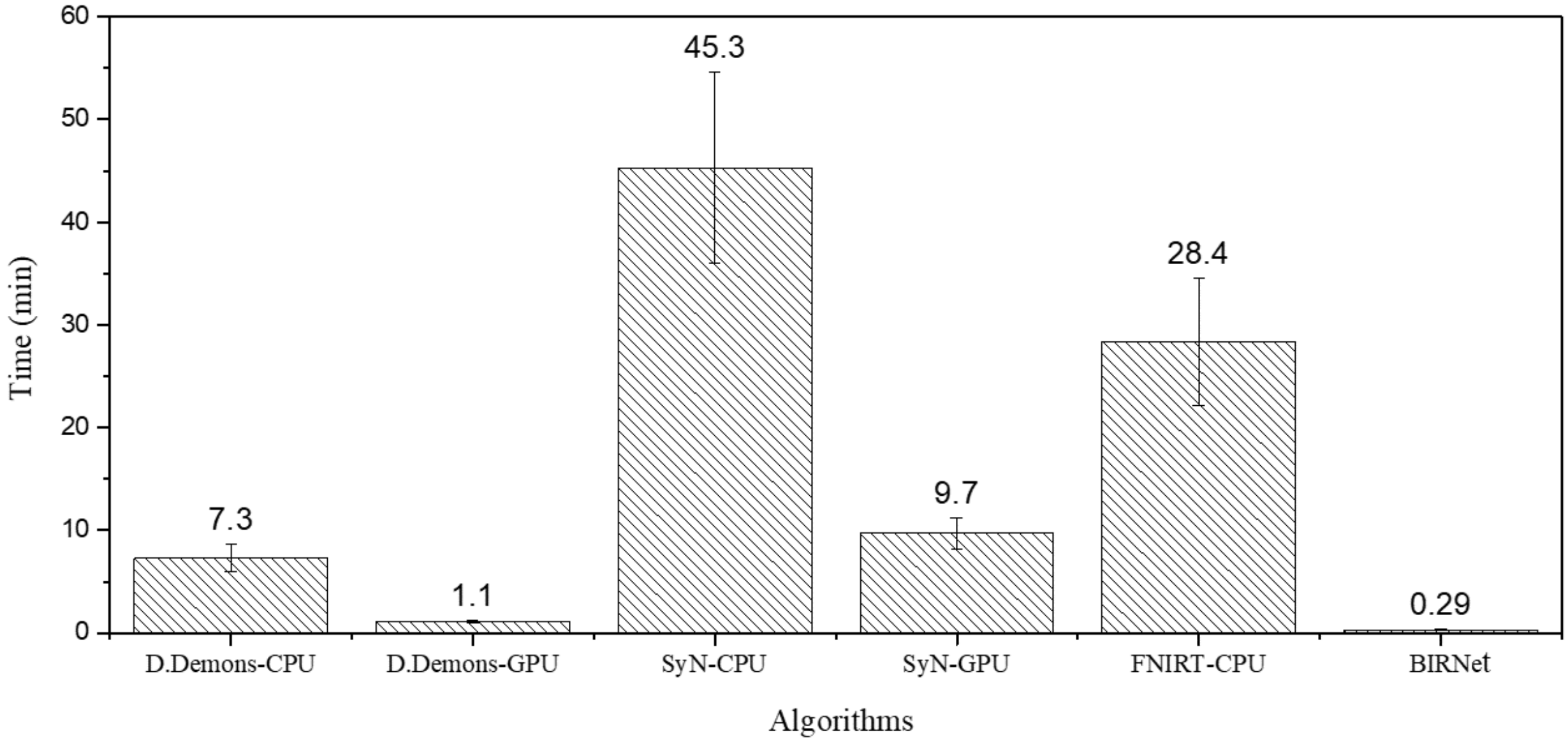

**Fig. 12.** Average computational times (in minutes) of different registration algorithms for registering a 220×220×184 brain image.

## 5. Conclusion

In this paper, we have introduced a dual-guided fully convolutional neural network, called BIRNet. To solve the issue of lacking ground-truth problem, BIRNet uses both pre-registered ground-truth deformation field and image similarity/difference metric to guide the training stage, thus making the deep learning model able to further refine the results. BIRNet employs strategies such as gap filling, hierarchical supervision, multi-channel inputs, and data augmentation for improving registration accuracy. Experimental results indicate that BIRNet achieves the state-of-the-art performance without the need for parameter tuning.

In summary, since the proposed BIRNet method is a general, fast, accurate, and easy-to-use method for brain image registration, it could be directly applied to many practical registration problems. However, there are still two issues that could be optimized in the future work. First, the recent model registers the subject image to a fixed template image, and thus we need to refine a simply transferred model for a new template image. Second, the smoothness of the predicted deformation field is supervised by the diffeomorphic training samples. Therefore, we will employ additional diffeomorphic constraint for the learning model to make the predicted deformation field exactly regular.

## Acknowledgment

This work was supported in part by NIH grants (EB006733, EB008374, MH100217, MH108914, AG041721, AG049371, AG042599, AG053867, EB022880, MH110274).

## References

Andersson, J.L., Jenkinson, M., Smith, S., 2007. Non-linear registration, aka Spatial normalisation FMRIB technical report TR07JA2. FMRIB Analysis Group of the University of Oxford 2.

Auzias, G., Colliot, O., Glaunes, J.A., Perrot, M., Mangin, J.-F., Trouve, A., Baillet, S., 2011. Diffeomorphic brain registration under exhaustive sulcal constraints. IEEE transactions on medical imaging 30, 1214-1227.

Avants, B.B., Epstein, C.L., Grossman, M., Gee, J.C., 2008. Symmetric diffeomorphic image registration with cross-correlation: evaluating automated labeling of elderly and neurodegenerative brain. Medical image analysis 12, 26-41.

Cao, T., Singh, N., Jojic, V., Niethammer, M., 2015. Semi-coupled dictionary learning for deformation prediction, Biomedical Imaging (ISBI), 2015 IEEE 12th International Symposium on. IEEE, pp. 691-694.

Cao, X., Yang, J., Zhang, J., Nie, D., Kim, M., Wang, Q., Shen, D., 2017. Deformable Image Registration Based on Similarity-Steered CNN Regression, International Conference on Medical Image Computing and Computer-Assisted Intervention. Springer, pp. 300-308.

Fan, J., Yang, J., Ai, D., Xia, L., Zhao, Y., Gao, X., Wang, Y., 2016a. Convex hull indexed Gaussian mixture model (CH-GMM) for 3D point set registration. Pattern Recognition 59, 126-141.

Fan, J., Yang, J., Lu, F., Ai, D., Zhao, Y., Wang, Y., 2016b. 3-Points Convex Hull Matching (3PCHM) for fast and robust point set registration. Neurocomputing 194, 227-240.

Fan, J., Yang, J., Zhao, Y., Ai, D., Liu, Y., Wang, G., Wang, Y., 2017. Convex Hull Aided Registration Method (CHARM). IEEE transactions on visualization and computer graphics 23, 2042-2055.

Fluck, O., Vetter, C., Wein, W., Kamen, A., Preim, B., Westermann, R., 2011. A survey of medical image registration on graphics hardware. Computer methods and programs in biomedicine 104, e45-e57.

Gutiérrez-Becker, B., Mateus, D., Peter, L., Navab, N., 2016. Learning optimization updates for multimodal registration, International Conference on Medical Image Computing and Computer-Assisted Intervention. Springer, pp. 19-27.

Gutierrez-Becker, B., Mateus, D., Peter, L., Navab, N., 2017. Guiding multimodal registration with learned optimization updates. Medical Image Analysis.

He, K., Zhang, X., Ren, S., Sun, J., 2015. Delving deep into rectifiers: Surpassing human-level performance on imagenet classification, Proceedings of the IEEE international conference on computer vision, pp. 1026-1034.

Ioffe, S., Szegedy, C., 2015. Batch normalization: Accelerating deep network training by reducing internal covariate shift, International Conference on Machine Learning, pp. 448-456.

Jia, Y., Shelhamer, E., Donahue, J., Karayev, S., Long, J., Girshick, R., Guadarrama, S., Darrell, T., 2014. Caffe: Convolutional architecture for fast feature embedding, Proceedings of the 22nd ACM international conference on Multimedia. ACM, pp. 675-678.

Johnson, H.J., Christensen, G.E., 2002. Consistent landmark and intensity-based image registration. IEEE transactions on medical imaging 21, 450-461.

Kim, M., Wu, G., Wang, Q., Lee, S.-W., Shen, D., 2015. Improved image registration by sparse patch-based deformation estimation. NeuroImage 105, 257-268.

Kim, M., Wu, G., Yap, P.-T., Shen, D., 2012. A general fast registration framework by learning deformation–appearance correlation. IEEE Transactions on Image Processing 21, 1823-1833.

Kingma, D., Ba, J., 2014. Adam: A method for stochastic optimization. arXiv preprint arXiv:1412.6980.

Klein, A., Andersson, J., Ardekani, B.A., Ashburner, J., Avants, B., Chiang, M.-C., Christensen, G.E., Collins, D.L., Gee, J., Hellier, P., 2009. Evaluation of 14 nonlinear deformation algorithms applied to human brain MRI registration. NeuroImage 46, 786-802.

Klein, S., Staring, M., Murphy, K., Viergever, M.A., Pluim, J.P., 2010. Elastix: a toolbox for intensity-based medical image registration. IEEE transactions on medical imaging 29, 196-205.

Loeckx, D., Maes, F., Vandermeulen, D., Suetens, P., 2003. Non-rigid image registration using a statistical spline deformation model, Biennial International Conference on Information Processing in Medical Imaging. Springer, pp. 463-474.

Luo, Y.-g., Liu, P., Shi, L., Luo, Y., Yi, L., Li, A., Qin, J., Heng, P.-A., Wang, D., 2015. Accelerating Neuroimage Registration through Parallel Computation of Similarity Metric. PloS one 10, e0136718.

Muyan-Ozcelik, P., Owens, J.D., Xia, J., Samant, S.S., 2008. Fast deformable registration on the GPU: A CUDA implementation of demons, Computational Sciences and Its Applications, 2008. ICCSA'08. International Conference on. IEEE, pp. 223-233.

Myronenko, A., Song, X., 2010. Intensity-based image registration by minimizing residual complexity. IEEE Transactions on Medical Imaging 29, 1882-1891.

Oliveira, F.P., Tavares, J.M.R., 2014. Medical image registration: a review. Computer methods in biomechanics and biomedical engineering 17, 73-93.

Ou, Y., Sotiras, A., Paragios, N., Davatzikos, C., 2011. DRAMMS: Deformable registration via attribute matching and mutual-saliency weighting. Medical image analysis 15, 622-639.

Rohé, M.-M., Datar, M., Heimann, T., Sermesant, M., Pennec, X., 2017. SVF-Net: Learning Deformable Image Registration Using Shape Matching, International Conference on Medical Image Computing and Computer-Assisted Intervention. Springer, pp. 266-274.

Ronneberger, O., Fischer, P., Brox, T., 2015. U-net: Convolutional networks for biomedical image segmentation, International Conference on Medical Image Computing and Computer-Assisted Intervention. Springer, pp. 234-241.

Rueckert, D., Frangi, A.F., Schnabel, J.A., 2001. Automatic construction of 3D statistical deformation models using non-rigid registration, International Conference on Medical Image Computing and Computer-Assisted Intervention. Springer, pp. 77-84.

Rueckert, D., Frangi, A.F., Schnabel, J.A., 2003. Automatic construction of 3-D statistical deformation models of the brain using nonrigid registration. IEEE transactions on medical imaging 22, 1014-1025.

Samant, S., Lee, S., Samant, S., 2015. GPU-Based Unimodal Deformable Image Registration in Radiation Therapy. Graphics Processing Unit-Based High Performance Computing in Radiation Therapy. Series: Series in Medical Physics and Biomedical Engineering, ISBN: 978-1-4822-4478-6. CRC Press, Edited by Xun Jia and Steve Jiang, pp. 129-148, 129-148.

Schmidhuber, J., 2015. Deep learning in neural networks: An overview. Neural networks 61, 85-117.

Serag, A., Aljabar, P., Ball, G., Counsell, S.J., Boardman, J.P., Rutherford, M.A., Edwards, A.D., Hajnal, J.V., Rueckert, D., 2012. Construction of a consistent high-definition spatio-temporal atlas of the developing brain using adaptive kernel regression. NeuroImage 59, 2255-2265.

Shamonin, D.P., Bron, E.E., Lelieveldt, B.P., Smits, M., Klein, S., Staring, M., Initiative, A.s.D.N., 2013. Fast parallel image registration on CPU and GPU for diagnostic classification of Alzheimer's disease. Frontiers in neuroinformatics 7.

Shattuck, D.W., Mirza, M., Adisetiyo, V., Hojatkashani, C., Salamon, G., Narr, K.L., Poldrack, R.A., Bilder, R.M., Toga, A.W., 2008. Construction of a 3D probabilistic atlas of human cortical structures. NeuroImage 39, 1064-1080.

Shen, D., Davatzikos, C., 2002. HAMMER: hierarchical attribute matching mechanism for elastic registration. IEEE transactions on medical imaging 21, 1421-1439.

Sokooti, H., de Vos, B., Berendsen, F., Lelieveldt, B.P., Išgum, I., Staring, M., 2017. Nonrigid Image Registration Using Multi-scale 3D Convolutional Neural Networks, International Conference on Medical Image Computing and Computer-Assisted Intervention. Springer, pp. 232-239.

Sotiras, A., Davatzikos, C., Paragios, N., 2013. Deformable medical image registration: A survey. IEEE transactions on medical imaging 32, 1153-1190.

ur Rehman, T., Haber, E., Pryor, G., Melonakos, J., Tannenbaum, A., 2009. 3D nonrigid registration via optimal mass transport on the GPU. Medical image analysis 13, 931-940.

Vercauteren, T., Pennec, X., Perchant, A., Ayache, N., 2009. Diffeomorphic demons: Efficient non-parametric image registration. NeuroImage 45, S61-S72.

Wang, Q., Kim, M., Shi, Y., Wu, G., Shen, D., Initiative, A.s.D.N., 2015. Predict brain MR image registration via sparse learning of appearance and transformation. Medical image analysis 20, 61-75.

Wu, G., Kim, M., Wang, Q., Shen, D., 2014. S‐HAMMER: Hierarchical attribute‐guided, symmetric diffeomorphic registration for MR brain images. Human brain mapping 35, 1044-1060.

Wu, G., Yap, P.-T., Kim, M., Shen, D., 2010. TPS-HAMMER: Improving HAMMER registration algorithm by soft correspondence matching and thin-plate splines based deformation interpolation. NeuroImage 49, 2225-2233.

Yang, X., Kwitt, R., Styner, M., Niethammer, M., 2017. Quicksilver: Fast predictive image registration–A deep learning approach. NeuroImage.